\documentclass{article}
\usepackage{ijcai19}

\usepackage{times}
\usepackage{soul}
\usepackage{url}
\usepackage[hidelinks]{hyperref}
\usepackage[utf8]{inputenc}
\usepackage[small]{caption}
\usepackage{graphicx}

\usepackage[textwidth=1.5cm]{todonotes}

\title{Using SMT Solvers to Validate Models for AI Problems}

\author{
Andrei Arusoaie$^1$,
Ionu{\c t} Pistol$^1$,
\\ 
$^1$ Alexandru Ioan Cuza University of Ia{\c s}i\\
andrei.arusoaie@uaic.ro,
ipistol@info.uaic.ro
}

\begin{document}

\maketitle

\begin{abstract}
Artificial Intelligence problems, ranging form planning/scheduling up to game
control, include an essential crucial step: describing a model which
accurately defines the problem's required data, requirements, allowed
transitions and established goals. The ways in which a model can fail are
numerous and often lead to a failure of search strategies to provide a quick,
optimal, or even any solution. This paper proposes using SMT (Satisfiability
Modulo Theories) solvers, such as Z3, to check the validity of a model. We
propose two tests: checking whether a final(goal) state exists in the model's described problem space and checking whether the transitions described can provide a path from
the identified initial states to any the goal states (meaning a solution has
been found). The advantage of using an SMT solver for AI model checking is
that they substitute actual search strategies and they work over an abstract
representation of the model, that is, a set of logical formulas. Reasoning at
an abstract level is not as expensive as exploring the entire solution space.
SMT solvers use efficient decision procedures which provide proofs for the
logical formulas corresponding to the AI model. A recent addition to Z3
allowed us to describe sequences of transitions as a recursive function, thus
we can check if a solution can be found in the defined model.
\end{abstract}

\section{Introduction}

An AI problem is generally solved by defining a model equivalent to a production system, with one or more initial states, one ore more goal (final) states and some transitions rules, and by providing a search strategy which is able to find a path between an initial state and a final state within the problem space described by the model.

From the STRIPS ~\cite{fikes1971strips} to more recent efforts ~\cite{russell2016artificial} and ~\cite{hopgood2016intelligent}, various methods to formalize AI problems (described in natural language) have been made. Although significant variations exists, a general procedure has been cemented, as described initially in ~\cite{amarel1981representations} and further detailed in ~\cite{russell2016artificial}. We follow four steps: we find a representation for a state of the problem, we identify initial and goal (final) states, we described the available transitions (and means to validate them) and we apply a search strategy within the problem space which is able to find a solution as a path between an initial and a final state. The resulting transition system can, however, be difficult to check, since the problem can hide many details (assumed data, hidden references), can be incorrectly interpreted and can have solutions difficult (or impossible) to recover. Checking whether a model can solve an AI problem has been a task considered since the original STRIPS language, with solutions ranging from mathematical demonstrations ~\cite{bylander1994computational} to computer aided efforts ~\cite{huang2014symbolic} and ~\cite{edelkamp2001mips}. In ~\cite{clarke2001bounded}, bounded model checking is tested in real scenarios with good results, which was one of the motivations for the effort presented in this paper.

A model can allow a solution (path) to be discovered, but it can also fail to do so, even if the problem has a solution. The reasons why a model can fail are varied, starting from an insufficiently detailed representation for a state, not including all required validations and transitions, even up to not considering particular search strategies which may be unable to find a solution even if it exists. 
We propose validating a model by checking whether two claims (implemented as logical formulas) can be proven as valid:
\begin{itemize}
    \item Valid final state (VFS): a valid state exists which checks the conditions for a final state. This would mean that a final state exists within the problem space, as defined by the model.
    \item Path to a final state (PFS): using valid transitions, as described in the model, a path can be found between the initial state and a final state.
\end{itemize}

If both claims are proven, a search strategy can be used to recover one or more solutions using the tested model. Checking PFS has the added benefit of providing such a solution, which could be used to choose or build a better search heuristic for this particular problem and model.

Indeed, checking these properties is not an easy task. The usual way to tackle this is to explore the solution space of the model using various strategies. In this paper we propose a different methodology for checking the desired properties of AI models. First, we encode the model  using \emph{logical formulas}, and thus, we keep an \emph{abstract} representation of the solution space. Then, we use an existing tool called Z3~\cite{DBLP:conf/tacas/MouraB08} which is able to \emph{automatically} reason about logical formulas. 
Reasoning at an abstract level is not as expensive as exploring the entire solution space. 
Finally, we analyze the results returned by Z3.

\paragraph{Satisfiability Modulo Theories}
Satisfiability Modulo Theories, shorthanded as SMT, are proposed solution to the problem of finding whether a (classical first-order) logical formula is \emph{satisfiable} with respect to logical theories. More precisely, provided with a logical formula with variables, the task is to find an assignment for the variables that make the formula true.
Solving the SMT problem is useful when it is restricted to logical theories: actual tools can be implemented to search the solutions of the problem. 
Such tools are called \emph{SMT solvers} and some of the more well-known are: Z3~\cite{DBLP:conf/tacas/MouraB08}, CVC4~\cite{cvc4} and AltErgo~\cite{alt-ergo}. 

The list of theories and procedures that SMT solvers are based on is extensive and includes simple linear arithmetic~\cite{presburger}, array theory~\cite{arrays}, uninterpreted functions to model-based theory combination~\cite{combination}, model based quantifier instantiation~\cite{instantiation}, simplex (linear real arithmetic)~\cite{simplex}, extensional array theory~\cite{extensional-array} and SAT abstraction/refinement~\cite{refinement}. 

To give the reader a glimpse of how SMT solvers work, we provide a classic example~\cite{DBLP:conf/sbmf/MouraB09} of a logical formula that involves integer arithmetic, arrays (encoded using the $\mathit{select}$ and $\mathit{store}$ axioms), and uninterpreted functions:
$$j = i + 2 \land f(\mathit{select}(\mathit{store}(a, i, 3), j - 2)) \neq f(j + 1 - i)$$

\noindent
Here, $\mathit{store}(a, i, 3)$ returns the array $a$ which stores value $3$ at position $i$, and $\mathit{select}(a, i)$ returns the value stored in $a$ at position $i$. The formula also contains an uninterpreted function $f$ (we are not aware of what $f$ computes).

First, an SMT solver performs a substitution of $j$ inside the formula, and thus, the first equality dissapears:
$$f(\mathit{select}(\mathit{store}(a, i, 3), i + 2 - 2)) \neq f(i + 2 + 1 - i)$$
Second, some simple arithmetic is performed:
$$f(\mathit{select}(\mathit{store}(a, i, 3), i)) \neq f(3)$$
Then, the SMT solver uses an axiom about the $\mathit{select}$ and $\mathit{store}$, which is included in a theory of arrays:
$$\mathit{select}(\mathit{store}(a, i, v), i) = v,$$
Therefore, after using this axiom, we obtain: $$f(3) \neq f(3).$$
Indeed, this formula is not satisfiable for any choice of the function $f$.

What is important to mention here is that the reasoning that Z3 performs is based on theories which guarantee that the results obtained by Z3 are formally correct. By encoding VFS and PFS in Z3 as logical formulas and checking their satisfiability, we obtain formal guarantees of correctness. 

For a given input formula, say $\varphi$, an SMT solver has three possible outputs: 
\begin{itemize}
    \item {\tt sat}: when the input $\varphi$ is \emph{satisfiable};
    \item {\tt unsat}: when the input $\varphi$ is not \emph{satisfiable};
    \item {\tt unknown}: $\varphi$ might be {\tt sat} or {\tt unsat}, but determining this is beyond the capabilities of the decision procedures.
\end{itemize}

SMT solvers scale in practice. For instance, Z3 is used at Microsoft for test case generation (e.g., PEX~\cite{pex}), program verification (e.g., Spec\#~\cite{spec}), static driver verification (e.g., SLAM~\cite{slam}).
Our choice for the Z3 SMT solver is motivated by several reasons: Z3 is used at an industrial scale; Z3 has native support for recursive functions (a quite recent addition to Z3, since November 20, 2018); and Z3 can be trusted: when it answers {\tt sat} then it produces a witness (an assignment for variables) that make the formula satisfiable; when it answers {\tt unsat} it produces a proof that the formula is not satisfiable.  
\subsection{Contributions.}
We propose a systematic approach to tackle the verification of the VFS and PFS properties for AI models.
First, we show how to encode these properties as logical formulas. Then, we explain how to use these encodings inside the Z3 SMT solver in order to obtain useful answers from it.
Next, we exemplify our approach on a model for a well-known AI toy problem and we provide an implementation in Z3 which is also available online. Finally, we analyze the results that we get when running the implemented Z3 code.

\subsection{Paper organization.}
Section 2 of this paper shows how the two tests (VFS and PFS) are defined as logical formulas and implemented in an SMT solver. Section 3 exemplifies the proposed method on a model for a classic AI problem and the results of testing the corresponding formulas in Z3, as well as the impact some changes have on the obtained results. Section 4 concludes the above sections and talks about possible future developments.

\section{Encoding PFS and VFS as logical formulas.}

We start by explaining our notations. 
We denote by $V(s)$ the predicate which is true when the state $s$ is valid, and by $F(s)$ the predicate which is true when the state $s$ is final. 
Transition functions are denoted by $t: \mathit{State} \times \mathit{Params} \rightarrow \mathit{State}$, and they receive as arguments a state and some parameters, and return a state. 
Indeed, the passed parameters should produce a valid state.
Multiple applications of a transition function to a state is denoted by $t^n(s, P)$, where $n$ is the number of applications and $P$ is a generic list of parameters: $$t^n(s, P) = t^{n-1}(t(s, \mathit{head}(P)), \mathit{tail}(P)).$$ 

\paragraph{Encoding VFS.} By checking this property we essentially check the model with respect to predicates valid ($V$) and final ($F$). If there is no state which is both valid and final then a final state could never be reached, since it doesn't exists within the searcheable problem space. 

In SMT, we analyse VFS by checking the satisfiability of the quantified conjunction: $\exists s. V(s) \land F(s)$.
If the solver returns {\tt sat} then it means that our model is consistent and it includes at least one final state in the problem space.
On the other hand, if the SMT solver returns {\tt unsat} then there is no point of implementing any transitions or search strategies, as no path to a goal state could exist in the current model.
If the solver returns {\tt unknown}, then the only information we have is that the procedures implemented by the solver were not sufficient to decide satisfiability. This usually means that finding a valid final state is very difficult, even if it exists, a difficulty which transfers to an eventual working implementation of that model. 
\paragraph{Encoding PFS.}
Encoding PFS as a logical formula is a little bit more complex. 
The problem is that we have to check if there exists a sequence of applications of the transition function (checking also the validity of the transition for its parameters) which starts with an initial state and ends with a final state. 
If we assume that $I(s)$ is the predicate which is true when $s$ is an initial state, then PFS is equivalent to the following formula:
$$\exists s.\exists n.\exists P. (I(s) \land F(t^n(s, P)))$$
Basically, the formula above says that there exists an initial state $s$ and there is a sequence of $n$ transitions which produces a final state starting with $s$. Moreover, the transitions to the final state are given by the list of parameters $P$.

When Z3 answers {\tt unsat} for this formula, then there is no initial state $s$ such that a final state can be reached from $s$ in the model. This information is a quite significant, since it tells us that, whatever strategy we use, we will never reach a final state.

On the other hand, if Z3 answers {\tt sat} then there is certainly a path to a final state starting from an initial state. Therefore, we are sure that the model allows us to find such paths. However, there is no guarantee that a particular strategy will recover solutions within this model. But Z3 also provides a solution as a set of values for witch the checked formula is satisfiable, including the list of parameters P. Having an example of a solution will allow an informed choice with regards to a search strategy. Also, we can adjust the problem's input parameters by adding various constraints over them and then use Z3 to look for solutions for particular instances. We discuss this further in Section~\ref{sec:smtlib}.

\paragraph{The SMTLib language.}
The above properties can be encoded into a language called SMT-LIB~\cite{BarFT-RR-17} which is accepted by the most important SMT solvers.
This language is meant to describe logical theories and logical formulas which are meant to be checked for satisfiability. The language has a syntax very similar to \emph{S-expressions} or Lisp.
Examples of SMT-LIB code is shown in Section~\ref{sec:smtlib}.

\section{Missionaries and Cannibals}

\subsection{Problem description}
\label{sec:problem}
This classic AI problem is well known, its generalization can be formulated as: On the shore of a river there are {\it nm} missionaries and {\it nc} cannibals. There is a boat with {\it bcap} capacity on the same shore. Find, if it exists, a way to move all people on the initial shore to the other shore, using the boat. Consider that the boat moves only with 1 to {\it bcap} people in it, and on neither shore there can be more cannibals than missionaries, if there is at least one missionary there.
\subsection{Model \#1: The valid variant}
\label{sec:modelcorrect}
A first example of a possible model for this problem will be described below. The model is an adaptation of the one proposed in \cite{amarel1981representations}. Consider a state of the problem a list of 6 values: $\mathit{\langle bcap, {nm}_1, {nc}_1, bp, {nm}_2, {nc}_2 \rangle }$, where:
\begin{itemize}
    \item $\mathit {bcap}$ is the capacity of the boat (maximum number of passengers)
    \item $\mathit{nm}_1$ and $\mathit{nc}_1$ are the numbers of missionaries and cannibals on the left (initial) shore
    \item $\mathit{bp}$ is the current location of the boat (either shore 1 or 2)
    \item $\mathit{nm}_2$ and $\mathit{nc}_2$ are the numbers of missionaries and cannibals on the right (goal) shore
\end{itemize}
Considering the problem statement in Section~\ref{sec:problem}, the following have to be true for a state to be valid (within the problem space):
\begin{itemize}
    \item $\mathit{nm}_1 + \mathit{nm}_2 = \mathit{nm}$
    \item $\mathit{nc}_1 + \mathit{nc}_2 = \mathit{nc}$
    \item $\mathit{bp}\in\big\{1,2\big\}$
    \item $\mathit{nm}_1>0 \rightarrow {nm}_1\geq {nc}_1$
    \item $\mathit{nm}_2>0 \rightarrow {nm}_2\geq {nc}_2$
\end{itemize}
The initial state would be $\mathit{\langle {bcap}, {nm}, {nc}, 1, 0, 0 \rangle}$, and the final state $\mathit{\langle {bcap}, 0, 0, 2, {nm}, {nc} \rangle}$.

A transition is configured by two parameters, $\mathit{mm}$ - the number of missionaries to be moved and $\mathit{mc}$ - the number of cannibals to be moved. The only transition available is\\

$t\Big{(}\langle bcap, nm_1, nc_1, bp, nm_2, nc_2 \rangle , mm, mc\Big{)} =$\\
\hspace*{4cm}$\langle bcap, nm_1', nc_1', bp', nm_2', nc_2' \rangle,$\\

which is valid if:
\begin{itemize}
    \item $0 < \mathit{mm}+{mc}\le{bcap}$
    \item $\mathit{bp}=1 \rightarrow {nm}_1'={nm}_1-{mm}$ and $\mathit{nc}_1'={nc}_1-{mc}$ and $\mathit{nc}_2'={nc}_2+{mc}$ and $\mathit{nc}_2'={nc}_2+{mc}$
    \item $\mathit{bp}=2 \rightarrow {nm}_1'={nm}_1+{mm}$ and $\mathit{nc}_1'={nc}_1+{mc}$ and $\mathit{nc}_2'={nc}_2-{mc}$ and $\mathit{nc}_2'={nc}_2-{mc}$
    \item $\mathit{bp'}=3-{bp}$
\end{itemize}
A solution (sequence of transitions) can be described as a list $\mathit{P}=\big\{\langle {mm}_{1}, {mc}_{1}\rangle, \langle {mm}_{2}, {mc}_{2}\rangle,...,\langle {mm}_{k}, {mc}_{k}\rangle\big\}$, where each element denotes a valid transition, first one from the initial state, last one to the final state.  
\paragraph{The model in SMT-LIB.}
\label{sec:smtlib}
The model discussed in Section~\ref{sec:modelcorrect} can be easily encoded as an SMT-LIB specification. For precision, we use version 2.0 of this language. 
Recall that SMT-LIB is a language accepted by many SMT solvers, including Z3. 

Let us recall the VFS and PFS properties: $\exists s . V(s) \land F(s)$ and  $\exists s.\exists n.\exists P. (I(s) \land F(t^n(s, P)))$. 
Encoding these existentially quantified formulas in SMT-LIB is done as follows: 
\begin{itemize}
    \item $s$, $n$ and $P$ are declared as uninterpreted functions/constants of their corresponding type:\\
    
    {\small {\tt 
(declare-const state State)\\
(declare-const n Int)\\
(declare-const p Parameters)\\
}}

    It is worth noting that {\tt \small  State} and {\tt\small Parameters} are just aliases for arrays of integers.

    \item The input parameters are uninterpreted constants too:\\
    
    {\small
    \tt
    (declare-const nm Int)\\
(declare-const nc Int)\\
    }

    \item The predicates $V$, $F$, and $I$ are implemented using functions that return a boolean value; here, we show only $F$:\\
    
    {\small 
    {\tt 
    (define-fun final ((s State)) Bool\\
  (and\\
\hspace*{5ex}(and\\
\hspace*{10ex}(= (nm2 s) nm)\\
\hspace*{10ex}(= (nc2 s) nc))\\
\hspace*{5ex}(= 2 (bp s))))\\
    }}
    
We use additional helper functions that extract information from the state: the call {\tt (bp s)} returns the boat position in the current state {\tt s}, {\tt (nm2 s)} returns the number of missionaries on shore 2, and {\tt (nc2 s)} returns the number of missionaries on shore 2. Therefore, a state is final if all the missionaries and cannibals are on shore 2.
    
The code corresponding to $I$ is available at \url{https://github.com/andreiarusoaie/z3-ai-model-verification/blob/master/experiments/cannibals_and_missionaries/correct-model/pfs.smt2#L65}, while the code corresponding to $V$ is available at \url{https://github.com/andreiarusoaie/z3-ai-model-verification/blob/master/experiments/cannibals_and_missionaries/correct-model/pfs.smt2#L36}.
     
    \item The function $t$ is defined as an ordinary function which takes a state and some parameters (e.g., missionaries and cannibals to be moved) and returns a new state:\\
    
    {\small
    {\tt
    (define-fun transition\\ 
    \hspace*{15ex}{\bf (}(s State) (mm Int) (mc Int){\bf )}\\ 
    \hspace*{6ex}State\\
    \hspace*{5ex}<function body>\\
    )\\
    }
    }
    The implementation in SMT-LIB for the transition function is available at \url{https://github.com/andreiarusoaie/z3-ai-model-verification/blob/master/experiments/cannibals_and_missionaries/correct-model/pfs.smt2#L85}.
    \item The function $t^n$ is defined as a recursive function which repeatedly applies $t$:\\
    
    {\small
    {\tt 
    (define-fun-rec tran\\ 
    \hspace*{15ex}{\bf (}(n Int)\\
    \hspace*{17ex}(state State)\\
    \hspace*{17ex}(params Parameters) 
    \hspace*{17ex} <additional params> {\bf )}\\
    \hspace*{5ex}State\\
    \hspace*{5ex}<function body with recursive call>\\
    )\\
    }
 }
 
 The corresponding code is available here: \url{https://github.com/andreiarusoaie/z3-ai-model-verification/blob/master/experiments/cannibals_and_missionaries/correct-model/pfs.smt2#L129}.
 
    \item VFS is now encoded as a simple conjunction:\\
    
    {\small 
    {\tt (assert\\
    \hspace*{2ex} (and (valid state) (final state))\\
    )\\}
    }
    
    \item PFS is a little bit more complex:\\
    
    {\small 
    {\tt (assert\\
\hspace*{2ex}(and\\
\hspace*{4ex}(initial state)\\
\hspace*{4ex}(final (tran n state p state (* 2 n)))\\
\hspace*{2ex}))\\
}}

Here, $p = mm_1, mc_1, \ldots, mm_k, mc_k$ is a list of length $2n$  which corresponds to a solution sequence $\mathit{P}=\big\{\langle {mm}_{1}, {mc}_{1}\rangle, \langle {mm}_{2}, {mc}_{2}\rangle,\ldots,\langle {mm}_{k}, {mc}_{k}\rangle\big\}$. 
The last two parameters are helper parameters that hold the last valid state and the size of $p$, respectively.
The task of the SMT solver is to find values for $n$, $s$, and $p$.
\end{itemize}

To check VFS or PFS we need to include the corresponding assertion in the implemented model. Consider that VFS and PFS can be satisfiable or not independent of the other.

In order to check our SMT-LIB specification for satisfiability we have to append the following command at the end of the specification:
{\small {\tt (check-sat)}}. 
If Z3 returns {\tt sat}, then it means that there are some values for $s$ (for VFS) or $s$, $n$ and $p$ (for PFS, respectively) that satisfy the specification. In order to obtain them we add {\small {\tt (get-model)}} right after {\small {\tt (check-sat)}}. 
If Z3 returns {\tt unsat}, then there are no values that satisfy the specification. 
This means that our model does not have the property we are currently checking and it might be something wrong with the model. 
If Z3 returns {\tt unsat} or {\tt unknown}, we can try to adjust the parameters of our model by adding various constraints (e.g., limit the boat capacity). 

In addition to all these, we can add problem specific restrictions to our specification.
For instance, we should add an assertion that limits the range of the parameters of the transition function, i.e., parameters should be between 0 and the boat size. This is acceptable, since exceeding the capacity of the boat in a transition will never lead us to a valid state.

\subsection{Other model variants}
\label{sec:alterations}

In order to see how Z3 can indicate failures in a model, we tried two alternatives to the model described above. 

\subsubsection{Model \#2}
First variation was to change these tests from the valid state checks: 
\begin{itemize}
    \item $\mathit {nm}_1>0 \rightarrow {nm}_1\geq {nc}_1$ changed to $\mathit{nm}_1\geq {nc}_1$
    \item $\mathit {nm}_2>0 \rightarrow {nm}_2\geq {nc}_2$ changed to $\mathit{nm}_2\geq {nc}_2$
\end{itemize}
This variation assumes that we check that the number of missionaries is larger than the number of cannibals while ignoring the requirement for this to be true only when the number of missionaries is non-zero. This is a likely assumption to make by a person developing the model when he/she is not experienced/attentive enough, especially since it requires less written code.

This change should make VFS invalid. Also, for most instances of this problem, a solution, unless $\mathit{bcap}$ is larger than $\mathit{nc}$, requires that, at least at one point, the number of missionaries on a shore be zero and the number of cannibals non-zero, thus we should also expect PFS to fail for most problem instances.

\subsubsection{Model \#3}
The second alternative model added an additional check to validate a transition: $\mathit{mm}>{mc}$. The requirement to always move more missionaries than cannibals could be a misinterpretation of the problem requirements that on all shores, if the number of missionaries is non-zero,  it is larger than the number of cannibals. Someone might assume that this requirement also applies to the boat.

This change would still allow a valid final state (thus VFS is satisfiable), but PFS should fail, as a solution for which always more missionaries than cannibals are moved shouldn't exist unless for particular instances in which $\mathit {nm} > {nc}$ and $\mathit {bcap} > 2$. 

\subsection{Analysis with Z3.}
The entire SMT-LIB specification discussed in Section~\ref{sec:modelcorrect} is available on Github: \url{https://github.com/andreiarusoaie/z3-ai-model-verification.git}.
Follow the accompanying instructions to run the code. Implementations for all variation discussed above and both VFS and PFS checks (performed separately) are included.

Typically, for every scenario that we created, we have two files -- one for each property: {\tt pfs.smt2} and {\tt vfs.smt2}. 
In every scenario, we constrained the boat size to be greater than 2, and we asserted that the number of missionaries and cannibals should be greater than 2 as well (to avoid trivial solutions):\\

{\tt
\noindent
(assert (< 2 missionaries))\\
(assert (< 2 cannibals))\\
(assert (< 2 (bcap state)))\\
}

For each scenario we present the elapsed time, the memory consumption, and the number of basic resource-consuming operations within the solver. 

We have performed the following tests on a machine with an Intel i7 8700K CPU at 3.7 GHz, and 32 GB of RAM. The system runs an Ubuntu based operating system called PopOS.

\paragraph{Tests for model \#1.} 
In the first scenario we used the model described in Section~\ref{sec:modelcorrect}. 
Z3 returns {\tt sat} for both PFS and VFS, indicating that each property is satisfiable, and thus proving that the model can describe at least one solution.
The results are summarized in Table~\ref{tbl:scenario1}. 

\begin{table}[h]
\centering
\begin{tabular}{|c|c|c|c|}
\hline
Scenario & Time & Memory & r-limit \\ 
\# 1 & (s) & (MB) & (no of ops.) \\ 
\hline
VFS &  0.01 & 2.9 & 2855 \\ 
\hline
PFS & 119.43 & 31.14 & 517650739 \\ 
\hline
\end{tabular}
\caption{Z3 statistics for model \#1}
\label{tbl:scenario1}
\end{table}

The {\tt (get-model)} command also gives us solutions. For VFS, Z3 finds the state $\langle 4,0,0,2,7722,3\rangle$, where the number of missionaries is 7722 and the number of cannibals is 3. Indeed, this state is final and valid. Note that for a different version of Z3 the returned solution could be different. 
A drawback for Z3 is the output which is not easy to read when it prints arrays or functions. This is common to all SMT solvers, and we recommend the use of an external tool to postprocess the output.

For PFS, Z3 returns {\tt sat} as well, and, when asked, it provides a solution. What is interesting to notice is that the number resource-consuming operations for PFS is much bigger than for VFS, and this is explained by the use of a recursive function. Solving the SMT problem when recursive functions are used is expensive, as pointed out by the values for memory consumption and elapsed time.

The almost two minutes time required to validate the existence of a sequence of recursive calls of the transition function is low, considering the unrestricted nature of the test. $\mathit {nm}$ and $\mathit {nc}$ are not restricted (except as being Integers and positive), $\mathit {bcap}$ is restricted to values between 2 and $\mathit {nm}$ and the parameters for the transitions ($\mathit {mm}$ and $\mathit {mc}$) are restricted between 0 and $\mathit {bcap}$. No other heuristics are used as to keep the results relevant for any possible search heuristic applied on the model.  
\paragraph{Tests for model \#2.}
In the second test we reproduce the first failure discussed in Section~\ref{sec:alterations}.
The error that we introduce in the model is meant to emphasize a common mistake: the lack of precision, i.e., we only say that the number of missionaries is bigger than the number of cannibals.
This is not always true, since the problem constraints allow us to have zero missionaries on shores. This change makes both the VFS and PFS invalid. 
The corresponding code for both VFS and PFS can be found here: \url{https://github.com/andreiarusoaie/z3-ai-model-verification/tree/master/experiments/cannibals_and_missionaries/variation1-bad-valid-function}.
For this scenario, Z3 returns {\tt unsat} for both properties. 

\begin{table}[h]
\centering
\begin{tabular}{|c|c|c|c|}
\hline
Scenario & Time & Memory & r-limit \\ 
\# 1 & (s) & (MB) & (no of ops.) \\ 
\hline
VFS & 0.01 & 2.87 & 1667 \\ 
\hline
PFS & 1.43 & 9.5 & 6676831 \\ 
\hline
\end{tabular}
\caption{Z3 statistics for model \#2.}
\label{tbl:scenario2}
\end{table}

In Table~\ref{tbl:scenario2} we show the statistics for this test. Both VFS and PFS fail in this case. Z3 is slower for PFS again, but it is able to decide in a fairly reasonable amount of time that this alteration of the model is not feasible. Since Z3 is sound, the result indicates the fact that  we will never be able to discover a solution. Moreover, this applies to any possible strategy. This is a very powerful result, which can be obtained only by reasoning about logical formulas.

\paragraph{Tests for model \#3.}
Finally, in the third test we experiment the second failure that we discuss in Section~\ref{sec:alterations}. In this case, the transition function is enriched with a new constraint: the number of missionaries in the boat is always bigger than the number of cannibals. Also, we limited $\mathit{nm}$ as equal to $\mathit{nc}$. As explained in Section~\ref{sec:alterations}, for VFS Z3 should return {\tt sat} and it does, while for PFS, Z3 returns {\tt unsat}. 

The behavior of Z3 for VFS should be the same as in Table~\ref{tbl:scenario1}. The change affects only PFS, whose code is available here: \url{https://github.com/andreiarusoaie/z3-ai-model-verification/blob/master/experiments/cannibals_and_missionaries/variation2-bad-transition/pfs.smt2}.

For the statistics shown in Table~\ref{sec:alterations}, we added a  constraint for the depth of the recursion to be less than 100. Considering the number of variables (even the instance is not set, parameters for the transitions are not constant), the number of variations attempted at each step is significant, nearly reaching the maximum available RAM for our test system. 

\begin{table}[h]
\centering
\begin{tabular}{|c|c|c|c|}
\hline
Scenario & Time & Memory & r-limit \\ 
\# 1 & (s) & (MB) & (no of ops.) \\ 
\hline
VFS & 0.01 & 2.9 & 28555\\ 
\hline
PFS & 4854.07 & 545.91 & 103893362134 \\ 
\hline
\end{tabular}
\caption{Z3 statistics for model \#3.}
\label{tbl:scenario3}
\end{table}

The statistics shown in Table~\ref{tbl:scenario3} shows that the time spent by Z3 to decide unsatisfiability is much larger than for the satisfiable model \#1. For PFS, the intended error in the model is significantly harder to prove and Z3 needs to almost 2000 times more operations to reach this conclusion.

A positive sign is that for no test the result was "unknown", as that would assume the insufficiency of Z3's decision procedure for this model. In additional tests we found that "unknown" is concluded extremely rarely and was always a consequence of errors in the SMT implementation of the model. 
\subsubsection{Discussion}
Z3 allows various tricks and tweaks. For instance, not only that you can use it to find whether the AI model  has certain properties, but you can add various constraints to check more powerful properties. For instance, in the first scenario, one can search for solutions where $\mathit{bc}$, $\mathit{nm}$ and $\mathit{nc}$ are bigger than, less than, or equal to some specified value. Knowing that a model is valid can lead to checking the existence of solutions for particular instances, even discovering limits or correlations between the existence of a solution and the values of these parameters. Constraints set on the values in $\mathit{P}$ - for example considering that people only of a single type can use the boat at once - can allow for testing various other variations. 
By setting an upper bound or a lower bound for the recursion depth we can test the existence of solutions of length between those bounds.

More powerful tools for proving logical formulas are interactive theorem provers like Coq~\cite{Coq}, or Isabelle~\cite{Nipkow:2002:IPA:1791547}. Unfortunately, these are not automatic and require a lot of professional training. On the other hand, other SMT solvers like CVC4~\cite{cvc4} or AltErgo~\cite{alt-ergo}, can be used in parallel with Z3. Note that SMT solvers cannot contradict each other: it is impossible for one to return {\tt sat} and another one to return {\tt unsat} for the same input formula. What can be different is the amount of time spent to compute the result and the complexity of the implementation. Also, only Z3 has native support for recursive functions which are required to test PFS or similar properties.

\section{Conclusion and Future work}
We showed that the benefits of a state-of-the-art SMT solver used to validate transition systems (as models for AI problems) can be significant. The added value of SMT solvers consists in the fact that the entire model is encoded using logical formulas. These formulas allow reasoning at an abstract level, and thus, it decreases the search space. 
The very recent addition of support for recursive functions in Z3 allowed us to look for and recover solutions, a new capability for bounded model checking. 

The proposed application of an established SMT solver in AI model validations comes with the minimal overhead of having to describe that model as SMT logical inferences. This effort proved to be manageable, the main difficulty being maintaining the semantic accuracy of the transcription. A potential automated transcription between STRIPS and Z3 code is planned as a future development for our system. PDDL ~\cite{mcdermott1998pddl} or other action languages could be considered later on as well.

\bibliographystyle{named}
\bibliography{references}

\end{document}